\documentclass[a4paper,fleqn]{cas-dc}

\usepackage[numbers]{natbib}
\usepackage{amssymb}
\usepackage{lipsum}
\usepackage{booktabs}
\usepackage{makecell}

\usepackage{algorithm}
\usepackage{algpseudocode}
\usepackage{graphicx}
\usepackage{subcaption}
\usepackage{amsmath}
\usepackage{float}

\def\tsc#1{\csdef{#1}{\textsc{\lowercase{#1}}\xspace}}
\tsc{WGM}
\tsc{QE}
\tsc{EP}
\tsc{PMS}
\tsc{BEC}
\tsc{DE}

\begin{document}

\shorttitle{Efficient Coordination with the System-Level Shared State: An Embodied-AI Native Modular Framework}
\shortauthors{Deng et al.}

\title [mode=title]{Efficient Coordination with the System-Level Shared State: An Embodied-AI Native Modular Framework}

\author[1,3]{Yixuan Deng}[
    role=First Author,
]
\author[2]{Tongrun Wu} [
    role=co-First Author,
]
\author[3,5] {Donghao Wu} [
]

\author[6]{Zeyu Wei}[
]

\author[3]{Jiayuan Wang} [
]
\author[1,2,3]{Zhenglong Sun} [
]
\author[4]{Yuqing Tang} [
]

\author[1,2,3]{Xiaoqiang Ji}[
    role=Corresponding Author,
    orcid=0000-0002-8556-3579
]
\ead{jixiaoqiang@cuhk.edu.cn}

\credit{Conceptualization, Methodology, Software, Writing - Original Draft}

\address[1]{School of Science and Engineering, The Chinese University of Hong Kong, Shenzhen, 2001 Longxiang Boulevard, Shenzhen, China}

\address[2]{School of Artificial Intelligence, The Chinese University of Hong Kong, Shenzhen, 2001 Longxiang Boulevard, Shenzhen, China}

\address[3]{Shenzhen Institute of Artificial Intelligence and Robotics for Society, Shenzhen, China}

\address[4]{International Digital Economy Academy, Shenzhen-Hong Kong Collaborative Innovation Center, Shenzhen, China}

\address[5]{School of Data Science, The Chinese University of Hong Kong, Shenzhen, 2001 Longxiang Boulevard, Shenzhen, China}

\address[6]{The School of Computer Science, The University of Sydney, Sydney, Australia}

\cortext[cor1]{Corresponding author}

\begin{abstract}
As Embodied AI systems move from research prototypes to real world deployments, they tend to evolve rapidly while remaining reliable under workload changes and partial failures.
In practice, many deployments are only \emph{partially decoupled}: middleware moves messages, but shared context and feedback semantics are implicit, causing interface drift, cross-module interference, and brittle recovery at scale.
We present \textsc{ANCHOR}, a modular framework that makes decoupling and robustness explicit system-level primitives.
\textsc{ANCHOR} separates (i) \emph{Canonical Records}, an evolvable contract for the standardized shared state, from (ii) a \emph{communication bus} for many-to-many dissemination and feedback-oriented coordination, forming an inspectable end-to-end loop.
We validate closed-loop feasibility on a de-identified workflow instantiation, characterize latency distributions under varying payload sizes and publish rates, and demonstrate automatic stream resumption after hard crashes and restarts even with shared-memory loss.
Overall, \textsc{ANCHOR} turns ad-hoc integration glue into explicit contracts, enabling controlled degradation under load and self-healing recovery for scalable deployment of closed-loop AI systems.
\end{abstract}



\maketitle

\section{Introduction}
\label{Intro}

Embodied AI has progressed rapidly across autonomous driving, robotics, and aerial autonomy.
In driving, unified end-to-end frameworks increasingly model the full stack from perception to planning (e.g., UniAD, CVPR 2023)~\cite{hu2023planning};
in robotics, learning-based controllers that execute reliably on real hardware have advanced quickly (e.g., diffusion policy, RSS 2023)~\cite{chi2023diffusion};
and in aerial autonomy, highly dynamic real-world systems have reached (and in some settings rivaled) top human performance (e.g., the swift drone racing system, Nature 2023)~\cite{kaufmann2023champion}.
This fast-moving landscape places growing pressure on the \emph{system layer}: practical deployments must support rapid iteration, where heterogeneous components evolve and recombine, while still providing dependable runtime behavior as task demands expand.

To keep pace with iteration pressure, the system layer must provide \emph{decoupling} as a default engineering property.
In practice, embodied deployments rarely evolve as a cleanly separated pipeline; instead, perception, state maintenance, decision making, tool use, and execution control
co-evolve at different cadences and are repeatedly recomposed under changing tasks and hardware.
Decoupling matters because it allows components to be replaced independently, constrains how load spikes and faults propagate across modules, and reduces integration overhead.

However, many deployed stacks remain only \emph{partially decoupled}.
Even when publish--subscribe middleware is used, interface contracts are often implicit (e.g., undocumented message semantics, ad-hoc topic conventions, duplicated context),
and coordination logic tends to leak across module boundaries as the system grows.
As a result, iteration frequently induces interface drift and hidden coupling, and integration cost can dominate development cycles~\cite{macenski2022robot,eugster2003many}.
This motivates treating decoupling not merely as a middleware choice, but as a \emph{system-level design commitment}.

This motivation underlies the long-standing use of modular middleware and componentized architectures in robotics, ranging from earlier robot platforms and middleware
(e.g., YARP and Orocos)~\cite{metta2006yarp, bruyninckx2001open}
to lightweight robotics messaging systems (e.g., LCM)~\cite{huang2010lcm},
to modern deployments such as ROS~2~\cite{macenski2022robot},
as well as microservice-based control architectures that emphasize reuse and interruptible concurrency in safety-critical mobile robots~\cite{schrick2025microservice}.
These systems also reflect general publish--subscribe principles that decouple producers and consumers~\cite{eugster2003many}.
We therefore view decoupling-enabled modularity as an engineering amplifier: it does not prescribe a learning paradigm (end-to-end or otherwise), but it can substantially improve maintainability, evolvability, and robustness in deployment.

Beyond evolvability, real-world physical environments impose a second constraint: robustness under long-horizon operation and deployment-time disturbances.
In real-world physical environments, embodied agents must operate for long durations under imperfect sensing and actuation, while continuously interacting with humans, other agents, and changing environments.
These settings inevitably introduce distribution shifts, long-tail events, and uncertainty, raising the bar for system-level safety and robustness.
Accordingly, recent work increasingly treats safety and robustness as first-class goals, including unified views of runtime assurance and safety filters~\cite{hsu2023safety},
safe motion planning under uncertainty in traffic~\cite{lei2025safe},
safe and platform-aware navigation modeling in complex terrain~\cite{roth2025learned},
broader discussions of robustness challenges for autonomous vehicles~\cite{chen2025toward},
and scalable learning-based safety filters in practice~\cite{nguyen2025gameplay}.
However, integrating these ideas into a \emph{reusable and evolvable systems framework} remains an important engineering challenge as embodied AI scales to diverse deployments: interfaces drift as modules evolve, shared context becomes fragmented across components, and recovery and degradation behaviors are often not made explicit at the system level~\cite{macenski2022robot, eugster2003many}. In many deployments, recovery, degradation, and fault-containment behaviors are not made explicit at the system layer and instead emerge from ad-hoc glue code and local conventions.
This implicitness makes robustness fragile under scale: when components are tightly coupled through shared-but-undefined context, anomalies and load fluctuations can amplify across the stack~\cite{macenski2022robot,eugster2003many}.

Motivated by these requirements, we introduce \textbf{ANCHOR}, a modular framework for embodied AI systems that provides stable engineering support amid rapid iteration and real-world deployment.
ANCHOR does not assume an end-to-end or a staged realization; instead, it offers clear system abstractions and interface semantics along the sensing-to-execution chain, enabling key capabilities to be replaced, extended, and composed independently.
Concretely, ANCHOR organizes the system around two system-level, stable interfaces:
\emph{(i) Canonical Records}, a shared and evolvable representation of normalized observations and system context, and
\emph{(ii) Communication Bus} that supports many-to-many dissemination, concurrency-aware delivery, and feedback interactions for closed-loop execution~\cite{eugster2003many}.
By making shared state and inter-component communication explicit system mechanisms (rather than implicit, ad-hoc glue),
ANCHOR aims to provide clearer operational behavior under load, partial failures, and restarts, while keeping pace with evolving task demands and model capabilities.

The main contributions of this work are as follows:
\begin{enumerate}
  \item \textbf{A modular framework viewpoint for embodied deployments.}
  We formulate the systems-layer need for an evolvable framework that supports rapid iteration, heterogeneous component recomposition, and robust runtime operation in real deployments~\cite{macenski2022robot, eugster2003many}.

  \item \textbf{Canonical Records as an explicit shared-state interface.}
  We introduce \textit{canonical records} as a stable and evolvable interface for normalized observations and shared system context, intended to reduce interface drift and implicit coupling across components.

  \item \textbf{A high-concurrency inter-component communication bus.}
  We design and implement a high-concurrency communication mechanism based on multi-producer multi-consumer (MPMC) dissemination and concurrency governance, enabling multiple command sources, multiple executors, and multiple observers, and supporting feedback interactions for closed-loop execution~\cite{eugster2003many}.

  \item \textbf{Evidence of modularity and robustness in a closed-loop framework.}
  We provide empirical evidence that ANCHOR supports modular recomposition via explicit contracts, controlled degradation under load, and automatic recovery after hard crashes and restarts, validating the system-level properties targeted by our design.

\end{enumerate}

The rest of this paper is organized as follows.
Section~2 reviews related work.
Section~3 presents the design of ANCHOR and provides an overall system view.
Section~4 presents empirical evaluation results.
Section~5 analyzes key mechanisms and critical paths.
Section~6 discusses limitations and future directions.
Finally, Section~7 concludes the paper.

\section{Related Work}
\label{sec:relatedwork}

We review prior work from three perspectives that motivate ANCHOR:
(i) embodied system architectures and middleware, (ii) robustness under deployment constraints, and
(iii) extensibility for rapid iteration.
And we distinguish what is \emph{explicitly defined and supported as a system-level primitive}
from what is left to downstream integration choices or external extensions.

\subsection{Embodied AI system architectures}
When embodied AI systems transition from research prototypes to real deployments, their architectures are rarely rebuilt from scratch.
Instead, they typically build on mature middleware and tooling ecosystems, organizing perception, planning, and control as composable components
integrated through message-based communication.
This line of work primarily targets engineering concerns such as integration, reuse, and coordination, rather than prescribing whether higher-level
algorithms must be end-to-end or staged.

ROS provides a widely adopted communications layer and tooling ecosystem to connect functional modules across heterogeneous compute nodes~\cite{Quigley2009ROS},
and ROS~2 re-architects the stack toward scalability and deployment readiness with documented real applications~\cite{macenski2022robot}.
Beyond ROS, modern embodied architectures exhibit recurring trade-offs:
they promote modularity and reuse while balancing concurrency, real-time constraints, and production readiness against overall complexity and maintainability.
For example, XBot2 emphasizes multi-threaded, mixed real-time (RT) and non-RT execution within a single system,
together with hardware abstraction and pluggable components for cross-platform reuse~\cite{laurenzi2023xbot2}.
In parallel, software-architecture research systematizes how architecture-based self-adaptation and reconfiguration
address runtime uncertainty and environmental changes~\cite{alberts2025software}.
Despite these ecosystems, decoupling is often \emph{incomplete} at the system level.
Middleware enables message exchange, but higher-level interface contracts---such as the semantics and lifecycle of shared context, and how feedback is incorporated across components---are frequently left to downstream integration choices.
As systems scale, these implicit contracts become a common source of hidden coupling and unpredictable cross-module interactions.

Recent real-time systems studies further show that even within ROS~2-style middleware, communication paths and scheduling interactions can materially shape end-to-end timing behavior~\cite{luo2025ros2}.
Complementary modeling work highlights that inter-process communication introduces nontrivial delay components that require explicit reasoning~\cite{luo2023modeling}.
Moreover, predictability can be sensitive to cross-layer priority effects when multiple components contend for shared resources~\cite{kim2025cros}. Executor policies and shared-resource scheduling in multi-threaded ROS~2 can further alter end-to-end behavior, effectively becoming a hidden coupling point across otherwise modular components~\cite{al2024dynamic}.
Together, these observations suggest that concurrency governance and delivery semantics should be treated as first-order system concerns rather than incidental configuration.

As embodied tasks become longer-horizon, more collaborative, and faster-evolving, architectural concerns increasingly shift from merely ``wiring modules to run''
toward enabling concurrent coordination among components and continual sensing--adjustment during execution.
Surveys on integrating foundation/large language models into robotics systems similarly organize the landscape around component views
(e.g., communication, perception, planning, and control) and emphasize filtering/correction mechanisms for stable real-world execution~\cite{kim2024survey}.
Motivated by this shift and without presupposing any learning paradigm, ANCHOR adopts a modular-framework perspective:
clear component boundaries and communication organization serve as the substrate for concurrent coordination and feedback-driven interaction,
supporting rapid iteration and extensible capability growth~\cite{wang2024large}.

\begin{table*}[t]
\centering
\small
\setlength{\tabcolsep}{3.5pt}
\renewcommand{\arraystretch}{1.1}
\caption{Explicit design coverage of system-level mechanisms.
\checkmark: explicitly provided as a system-level primitive with defined semantics/interfaces; $\times$: not a primary system-level primitive (even if achievable via extensions).}
\label{tab:explicit-coverage}
\begin{tabular}{lccccc}
\toprule
\textbf{Work} &
\makecell{\textbf{MPMC}\\\textbf{pub--sub}} &
\makecell{\textbf{Feedback}\\\textbf{primitives}} &
\makecell{\textbf{QoS /}\\\textbf{backpressure}} &
\makecell{\textbf{Observability}\\\textbf{\& replay}} &
\makecell{\textbf{Canonical}\\\textbf{shared state}} \\
\midrule
ROS/ROS2 & \checkmark & \checkmark & \checkmark & \checkmark & $\times$ \\
Behavior Trees (pattern) & $\times$ & $\times$ & $\times$ & $\times$ & $\times$ \\
Microservice control architectures & $\times$ & \checkmark & \checkmark & \checkmark & $\times$ \\
\textbf{ANCHOR (this work)} & \checkmark & \checkmark & \checkmark & \checkmark & \checkmark \\
\bottomrule
\end{tabular}
\end{table*}

\subsection{Robustness under deployment constraints}
In real deployments, embodied systems must not only complete tasks but also maintain stable and safe execution under uncertainty and dynamic interactions,
making robustness and safety first-class requirements.
On the algorithmic side, runtime assurance and safety-filtering paradigms monitor and (when necessary) intervene on nominal policies during execution~\cite{hsu2023safety}.
On the systems and architecture side, recent efforts emphasize engineering mechanisms such as interruptibility, concurrency, and fail-safety,
and explore distributed organizational structures to mitigate brittleness from centralized logic~\cite{schrick2025microservice}.

Many approaches further introduce explicit and interpretable execution structures---including hierarchical state machines,
behavior trees, and crash-only recovery-style designs---to support online monitoring, rapid interruption, and error recovery,
while decomposing complex tasks into reusable units~\cite{harel1987statecharts, colledanchise2018behavior, candea2003crash}.
At the architecture level, organizing robot functionality into finer-grained services/components and reducing coupling to centralized logic
can improve concurrent coordination and contain fault propagation under anomalies~\cite{schrick2025microservice}.
Behavior trees are widely used in practice due to their modular conditions/actions and natural support for fallback and retries;
recent work also systematizes BT properties and evaluation metrics, highlighting the need for consistent measurements~\cite{gugliermo2024evaluating}.

These directions provide important foundations for robust execution via execution structure, monitoring, and recovery.
However, in real deployments robustness is often constrained by systems-layer coupling:
when many modules run concurrently, information fans out to multiple consumers, and feedback must be incorporated online, robustness depends on how coordination is realized across components.
Without an explicit shared representation of state and context, interface drift and implicit coupling can be amplified by iteration and load fluctuations, increasing fault propagation and debugging costs.

Recent analyses of executor behavior in ROS~2 highlight that runtime scheduling choices can measurably affect end-to-end responsiveness in component pipelines~\cite{tang2023real}.
Separately, timing analysis of processing chains with data refreshing shows that seemingly local design decisions can propagate into system-level latency and stability properties~\cite{tang2024timing}.
Together, these observations reinforce a trend toward making coordination, feedback, and fault-containment \emph{explicit} as system mechanisms rather than emergent behavior from glue code.
ANCHOR targets this systems-layer gap by aligning (i) a unified yet evolvable shared representation (\textit{canonical records}) and
(ii) concurrency-friendly inter-component communication and feedback as explicit system mechanisms.

\subsection{Extensibility for rapid iteration}
Rapid iteration in embodied systems often occurs at the level of capability composition:
sensing modalities are added or replaced, models and policies are updated frequently,
and execution stacks evolve with changing hardware form factors and task demands.
Deployment-oriented systems therefore must allow components to iterate at different cadences and minimize full-stack rewiring and interface refactoring;
otherwise, integration overhead quickly dominates development cycles~\cite{macenski2022robot}.

In practice, extensibility is accumulated through infrastructure at different layers.
Mixed RT/non-RT middleware emphasizes pluginized components and hardware abstraction to reduce coupling when integrating new devices and control modules~\cite{laurenzi2023xbot2},
while mature open-source  planning/\\manipulation stacks often rely on community-driven extensions to close capability gaps and evolve via incremental additions~\cite{fresnillo2023extending}.
As embodied systems move toward heterogeneous integration and multi-agent collaboration, extensibility becomes an organizational capability across platforms, vendors, and subsystems:
fleet and infrastructure management frameworks highlight standardized interfaces and \\ adapter-based integration to append new robots and devices without rewriting the entire system~\cite{valner2022scalable}.
Despite these building blocks, the literature consistently notes that rapid iteration remains constrained by interface drift, dependency growth, and runtime uncertainty;
consequently, architecture-level mechanisms for adaptation and evolvability remain actively studied yet not fully converged~\cite{alberts2025software}. Empirical comparisons also suggest that middleware backends can exhibit distinct latency and robustness behaviors under dynamic network conditions, turning backend selection and tuning into a recurring integration decision as systems scale~\cite{chovet2025performance}.

Recent work on workflow-oriented and context-aware orchestration in ROS-based deployments emphasizes that integration increasingly involves explicit workflow logic and shared context management,
rather than simple point-to-point connections~\cite{ochoa2024dynamic}.
Related frameworks for hybrid edge--cloud robot-assisted operations likewise highlight the practical need to manage shared context and task assignment across distributed resources~\cite{chauhan2024kriota}.
These trends amplify the value of stable interfaces for shared context and coordination as systems evolve and scale.
From this perspective, ANCHOR is positioned as a modular framework: it emphasizes clear component boundaries and an evolvable shared representation (\textit{Canonical Records}),
and uses concurrency-friendly inter-component communication to sustain multi-component coordination, turning repeated refactoring into controlled, incremental evolution.

\subsection{Operational definition of first-class mechanisms}
To make comparisons concrete (Table~\ref{tab:explicit-coverage}), we use \emph{first-class mechanism} in an operational sense:
a concern is first-class if it is \emph{explicitly defined and supported by the system layer} (with clear semantics, interfaces, and lifecycle),
rather than being left to ad-hoc downstream integration or achievable only via external extensions.
This definition distinguishes what a framework provides as a direct primitive from what may be possible but is not a primary design commitment.

To make this landscape concrete, Table~\ref{tab:explicit-coverage} summarizes whether representative systems and patterns
\emph{explicitly} define several modularity and robustness concerns as first-class mechanisms (per the above definition).
We emphasize that this checklist reflects \emph{explicit design emphasis} rather than absolute capability.

Table~\ref{tab:explicit-coverage} highlights that prior systems often provide strong support for communication ecosystems,
QoS and operational tooling, or architectural isolation at specific layers~\cite{Quigley2009ROS, macenski2022robot, schrick2025microservice}.
ANCHOR builds on these foundations and aims to further elevate a canonical shared-state contract (\textit{canonical records})
as an explicit system-level mechanism, aligning shared context with concurrency-friendly dissemination and feedback to support robust coordination under deployment constraints.

\section{ANCHOR Design}
\label{sec:design}

\subsection{System Overview}
ANCHOR is a modular framework for embodied AI systems designed to support rapid iteration and robust runtime operation.
Rather than committing to an end-to-end or staged realization, ANCHOR provides stable system-level abstractions and interface semantics along the sensing-to-execution chain,
so heterogeneous components can evolve independently while remaining composable in deployment.

At high level, ANCHOR organizes the system around two explicit primitives:
\textit{canonical records}, which provide a shared and evolvable representation for normalized observations and system context,
and a \emph{communication bus}, which provides many-to-many dissemination with concurrency-aware delivery and explicit support for feedback.
This separation makes both shared context and inter-component coordination explicit at the system layer, avoiding ad-hoc ``glue'' logic scattered across components.

\begin{figure*}[htb]
  \centering
  \includegraphics[width=\textwidth,height=0.4\textheight]{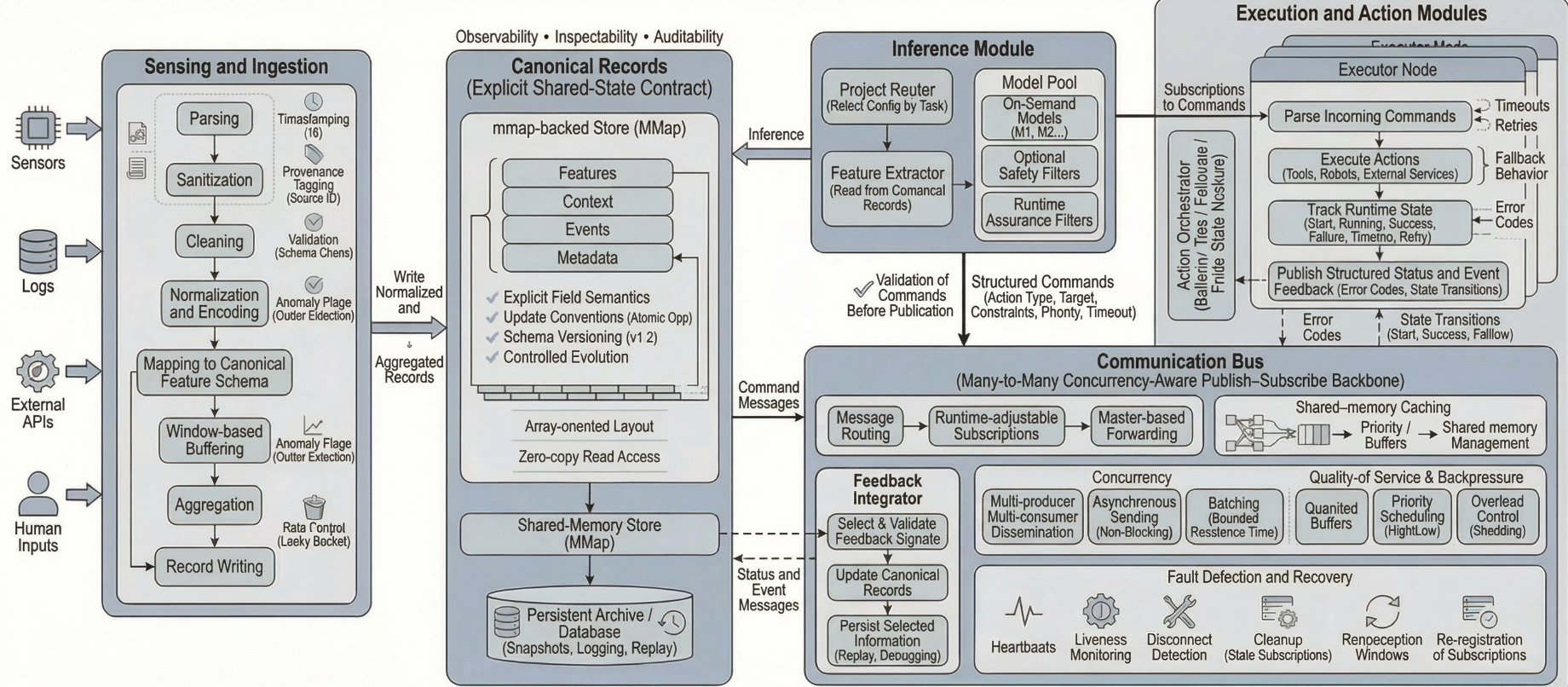}
  \caption{ANCHOR system overview. Upstream ingestion normalizes heterogeneous inputs into Canonical Records; inference consumes records and publishes commands/actions to the communication bus; execution subscribes and produces status/events; feedback is materialized back into Canonical Records, forming an explicit closed loop without a centralized monolithic controller.}
  \label{fig:anchor-overview}
\end{figure*}

A typical closed-loop runtime flow (Fig.~\ref{fig:anchor-overview}) proceeds as follows.
Upstream modules ingest raw signals and write normalized outputs into \textit{canonical records}.
Inference modules read the current context from \textit{canonical records} and publish action/control messages to the communication bus.
Execution modules subscribe to these messages, carry out corresponding actions, and publish status/events back to the bus.
Designated subscribers absorb feedback and materialize it as state updates in \textit{canonical records}, closing the loop while keeping component boundaries intact.
Because cross-component context and coordination are mediated by stable primitives, the system supports recomposition with reduced rewiring,
and its operational boundaries under concurrency can be reasoned about more directly.

\subsection{Design Goals}
ANCHOR is guided by two primary goals: \textit{modularity} and \textit{robustness under deployment constraints}.
We make these goals concrete in terms of what the system layer must provide.

\textbf{Modularity.}
Embodied deployments evolve quickly: sensors, models, and execution stacks change at different cadences.
The system layer should therefore provide stable interfaces that (i) localize changes within a component, (ii) minimize cross-module rewiring when components are added or replaced, and
(iii) reduce implicit coupling caused by undocumented message contracts or fragmented shared context.
ANCHOR pursues modularity by explicitly separating shared context (\textit{canonical records}) from coordination (\textit{communication bus}), and by defining clear semantics for both.

\textbf{Robustness.}
Real-world operation introduces uncertainty, load fluctuations, and partial failures.
Robustness in our setting is a property of runtime behavior under stress: the system should degrade in a controlled manner rather than exhibiting cascading failures,
and it should recover cleanly after transient disconnections or restarts.
ANCHOR pursues robustness through concurrency-friendly dissemination (MPMC), explicit feedback pathways, and runtime mechanisms such as bounded buffering, prioritization/QoS knobs,
and liveness monitoring for failure detection and recovery.

\subsection{Canonical Records}
\textit{Canonical records} provide an explicit shared-state interface for cross-component context.
The key motivation is to reduce interface drift and implicit coupling when multiple components must consume or update overlapping state.

\textbf{Standardized, evolvable representation.}
Instead of passing heterogeneous raw payloads through chains of point-to-point adapters, upstream modules normalize observations into a canonical schema.
This schema acts as a contract: fields have explicit semantics and update conventions, enabling independently developed components to interoperate with fewer hidden assumptions.
As the system evolves, the schema can be extended in a controlled manner, preserving compatibility where needed.

\textbf{Efficient sharing and persistence.}
\textit{Canonical records} are implemented on a shared memory region backed by \texttt{memmap}, enabling efficient read/write access across processes while also supporting persistence for logging and replay.
On top of \texttt{memmap}, ANCHOR adopts a uniform array-oriented layout contract, so heterogeneous components can access shared context without exposing internal implementation details.
This design reduces redundant copies and avoids forcing every interaction to go through the communication bus, while keeping state exchange explicit and inspectable at the system boundary.

\begin{figure*}[hbt]
  \centering
  \includegraphics[width=0.8\linewidth]{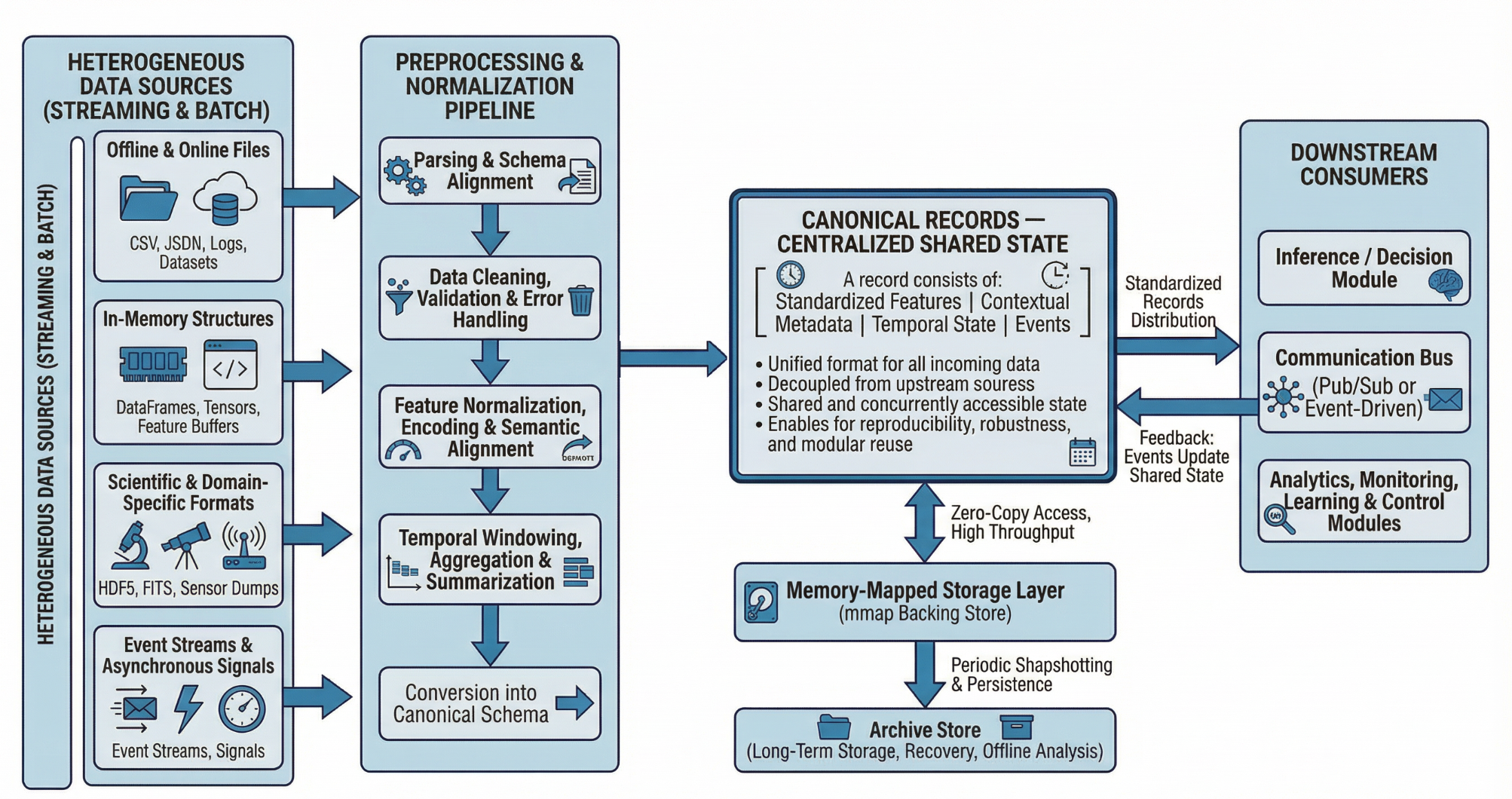}
  \caption{Canonical Records as an explicit shared-state contract. A memmap-backed region stores normalized observations and system context with an agreed layout and update conventions, enabling cross-process sharing, persistence, and controlled schema evolution.}
  \label{fig:canonical-records}
\end{figure*}

\textbf{Explicit responsibilities in closed-loop execution.}
\\ \textit{Canonical records} are where normalized context is accumulated and where feedback is materialized as state updates (Fig.~\ref{fig:canonical-records}).
Inference modules treat \textit{Canonical records} as read-only context for decision making, while write paths are explicitly associated with ingestion and feedback handlers.
This separation preserves clear responsibilities and simplifies debugging under concurrency.

\subsection{Communication Bus}
While \textit{Canonical records} anchor shared context, the communication bus anchors runtime coordination.
It adopts a publish--subscribe abstraction to decouple producers and consumers, and is designed to support many-to-many (MPMC) dissemination as a first-class pattern.

\textbf{Messages and interaction patterns.}
The bus carries two primary message classes:
(i) \emph{commands/actions} published by inference (or other decision-making) modules, and
(ii) \emph{status/events} published by execution modules.
Both message classes follow the same dissemination mechanism, which makes feedback an explicit part of the runtime interaction model rather than an afterthought.

\textbf{Roles and routing.}
To support both intra-cluster and cross-cluster coordination, the bus organizes routing into three roles:
\emph{nodes} (publishers/subscribers), a \emph{master} that maintains subscriptions and forwards messages within a cluster, and
\emph{gateways} that bridge traffic across clusters and re-inject messages into target clusters.
This structure preserves local publish--subscribe semantics while enabling scale-out beyond a single node group.

\begin{figure*}[htb]
  \centering
  \includegraphics[width=0.8\linewidth]{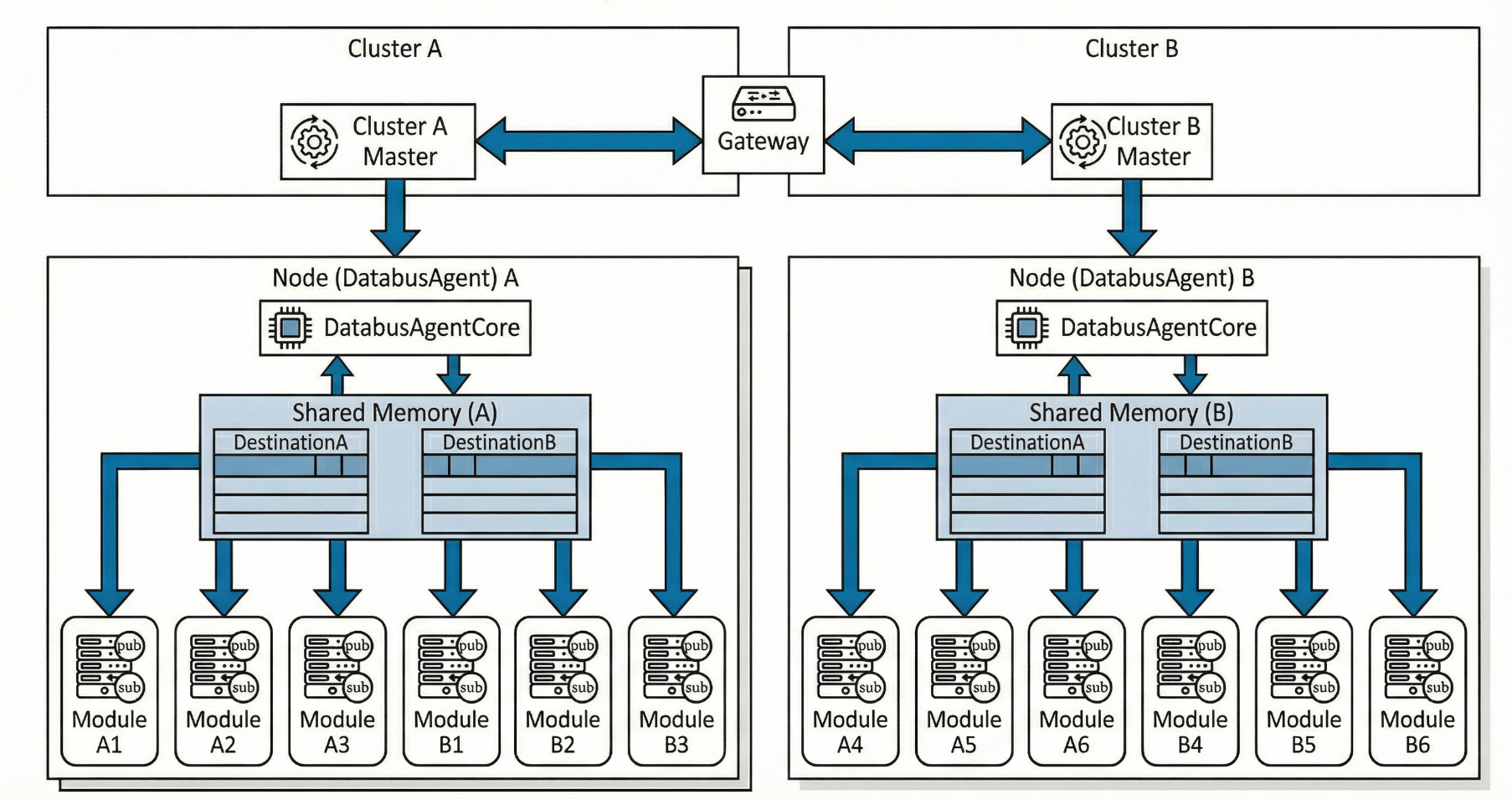}
  \caption{Communication bus roles and routing. A master maintains subscriptions and forwards messages within a cluster; gateways bridge cross-cluster traffic; nodes publish/subscribe to commands/actions and status/events for many-to-many coordination and feedback.}
  \label{fig:bus-routing}
\end{figure*}

\textbf{Topic structure and basic QoS.}
Topics follow a hierarchical structure such as \texttt{/channel/region/(nodeId)/prio}.
The \texttt{region} field controls dissemination scope (e.g., local vs cross-cluster),
\texttt{nodeId} supports directed delivery when needed, and
\texttt{prio} provides a basic priority signal that can be mapped from message types.
This organization helps keep addressing explicit and supports priority-aware handling under load.

\textbf{High-load handling.}
To reduce overhead under high throughput, nodes use shared-memory caches and asynchronous sending, and the bus supports batching for efficient transmission.
To avoid excessive delays at low traffic, a maximum residence time bounds how long messages may wait for batching.
Together, these mechanisms aim to keep delivery behavior controlled across a wide range of load regimes.

\textbf{Liveness and recovery.}
For transient disconnects and process failures, the bus uses liveness monitoring (e.g., heartbeats) to detect outages and trigger reconnection.
Reconnection includes re-registering identity and subscriptions so routes can be reconstructed without manual intervention.
These mechanisms provide a systems-layer basis for recovery and reduce the chance that a single component failure permanently breaks closed-loop operation.

\textbf{Runtime subscription management.}
In deployment, subscribers may need to adjust what they consume based on task phase and resource constraints.
The bus supports runtime subscription management so nodes can change subscriptions without global rewiring,
enabling the system to adapt component coordination under evolving workloads.

\subsection{Modularity and Robustness in Practice}
ANCHOR combines \textit{canonical records} and the communication bus to realize the design goals of \emph{modularity} and \emph{robustness} in practice.
\textit{Canonical records} stabilize cross-component context by making shared state explicit and evolvable, which localizes change and reduces interface drift as components iterate.
The communication bus complements this by governing concurrent coordination and feedback through explicit dissemination semantics, keeping interactions decoupled and inspectable at runtime.

From a modularity perspective, explicit shared context and publish--subscribe coordination reduce repeated integration work and global rewiring when components are added, replaced, or composed.
From a robustness perspective, bounded buffering and prioritization provide controlled behavior under overload, while liveness monitoring and reconnection logic enable clean recovery from transient failures or restarts.
We later characterize these runtime properties empirically through latency distributions and recovery behavior under load and failures.

\section{Empirical Evaluation}
\label{sec:eval}

\subsection{Overview and Scope}
\label{subsec:eval-overview}

This section empirically characterizes the runtime behavior of ANCHOR using an event-driven closed-loop automation workflow.
Multiple upstream producers ingest heterogeneous observation streams and periodically perform window-based preprocessing and aggregation, writing normalized state into \textit{canonical records} as a shared contract across components.
A policy inference module consumes the records and publishes action commands to a message bus; downstream executors subscribe to these commands, interact with remote service endpoints, and publish runtime status and outcome events back to the bus.
Selected feedback signals are integrated into \textit{canonical records} and archived for replay, enabling traceability across iterations.

Our goal is to validate end-to-end feasibility and characterize runtime properties of the messaging layer under concurrency and transient disruptions, rather than optimizing task-specific performance.
To avoid domain-specific disclosure, we abstract application details while preserving the system structure, interfaces, and representative workloads relevant to ANCHOR.

We evaluate ANCHOR along two axes.
First, we describe and validate a complete closed-loop execution pattern---from upstream aggregation into\textit{canonical records}, to command publication, to downstream execution and feedback materialization---to establish feasibility and traceability.
Second, we stress the message bus under increasing load and under transient disconnect/reconnect events, reporting distributional latency (ECDF and tail percentiles) and recovery behavior to characterize robustness at runtime.

\subsection{Implementation and Measurement Setup}
\label{subsec:eval-setup}

\textbf{Current testbed.}
All experiments are conducted on a single physical machine in our current test environment (CPU: AMD EPYC 9754 with 16 CPU cores available to the experimental environment; memory: 128\,GB DRAM; GPU: NVIDIA RTX 3090 with 24\,GB memory for inference-related runs; OS: Ubuntu 22.04).
The reported numbers should be interpreted as an empirical characterization under this testbed and the current implementation, rather than a hardware-agnostic performance bound.

\textbf{Workflow.}
The evaluation pipeline is instantiated as a set of independent component processes connected via \textit{canonical records} and \textit{message bus}.
Upstream producers ingest heterogeneous observation streams and perform periodic window-based aggregation, writing the aggregated state into \textit{canonical records}.
The policy module reads \textit{canonical records} and publishes action commands; one or more executor processes subscribe to these commands, invoke remote endpoints, and publish runtime status and outcome events.
For replay and debugging, commands and events are recorded during execution.

\textbf{Metrics.}
We report message-processing latency in a distributional form by plotting empirical cumulative distribution functions (ECDFs) and summarizing key percentiles (P50, P90, and P99).
For transient disconnect/reconnect tests, we additionally report detection time and reconnection time.
Unless otherwise stated, we reuse the same setup and measurement conventions throughout this section.

\subsection{Closed-Loop Execution Pattern}
\label{subsec:eval-closed-loop}

We validate a complete closed-loop execution pattern supported by ANCHOR.
The system proceeds in repeated cycles.
Upstream producers ingest raw inputs, parse and sanitize the inputs, normalize them into a unified representation, and aggregate them over a fixed window to obtain a shareable state; the state is then written into Canonical Records.
Algorithm~\ref{alg:preprocess} shows simplified pseudocode of the upstream preprocessing stage, where the key step is to map raw inputs into a canonical feature representation and update windowed aggregates in \textit{canonical records}.

\begin{figure*}[H]
\centering
\begin{subfigure}[H]{0.48\linewidth}
  \centering
  \includegraphics[width=0.7\linewidth]{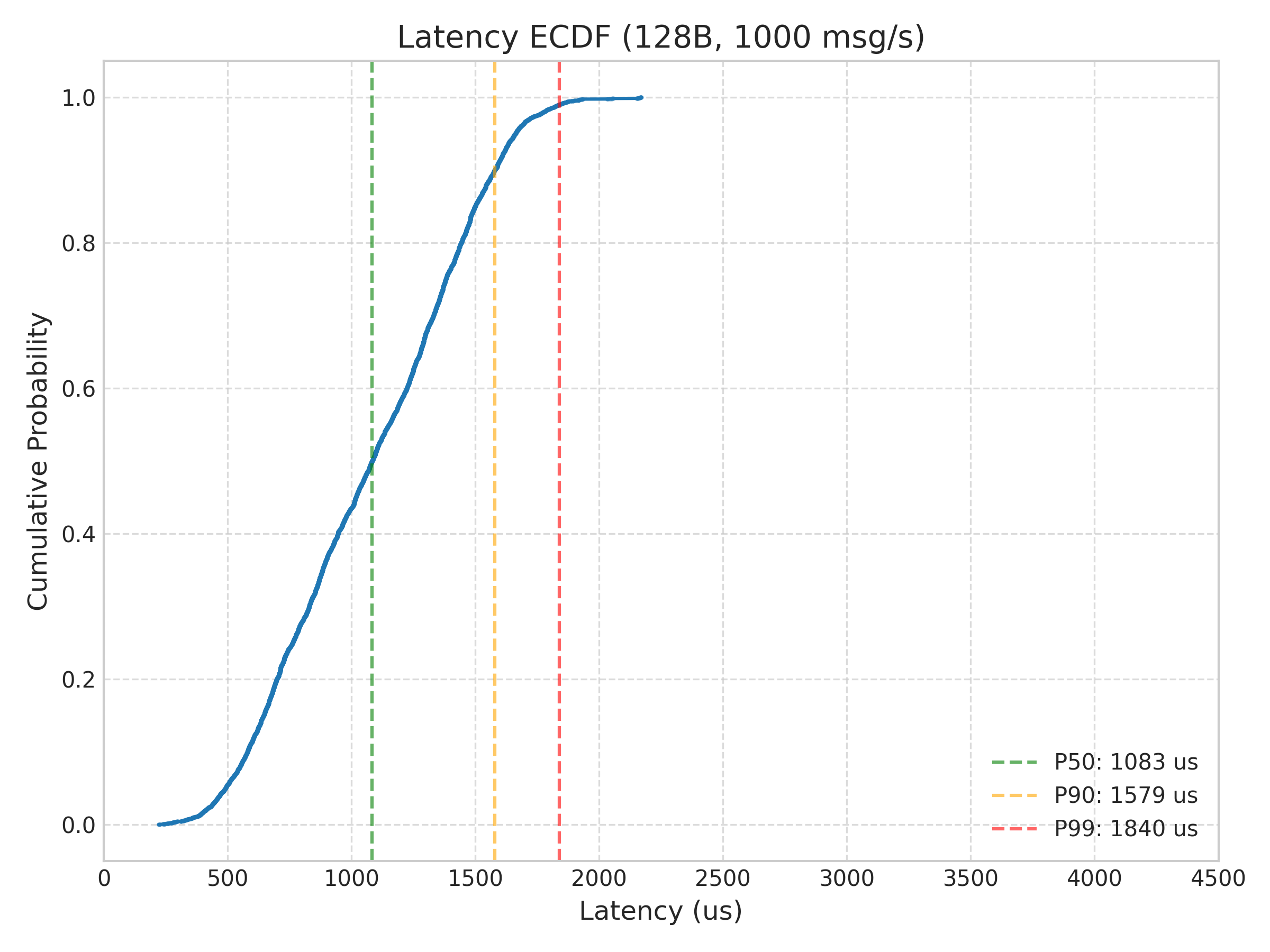}
  \caption{Payload=128B, Rate=1000 msg/s.}
  \label{fig:bus-ecdf-128-1k}
\end{subfigure}
\hfill
\begin{subfigure}[H]{0.48\linewidth}
  \centering
  \includegraphics[width=0.7\linewidth]{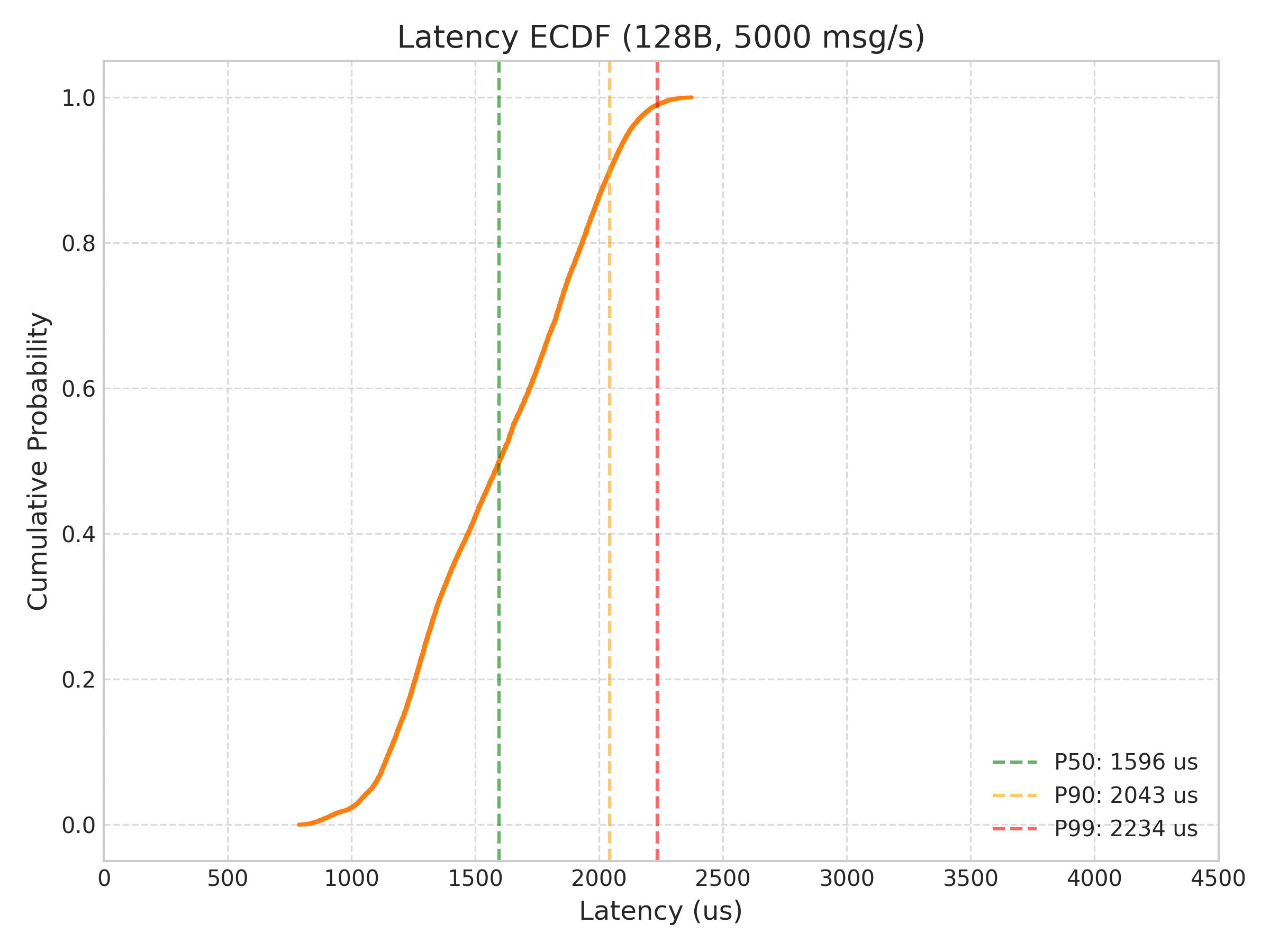}
  \caption{Payload=128B, Rate=5000 msg/s.}
  \label{fig:bus-ecdf-128-5k}
\end{subfigure}

\vspace{0.5em}

\begin{subfigure}[H]{0.48\linewidth}
  \centering
  \includegraphics[width=0.7\linewidth]{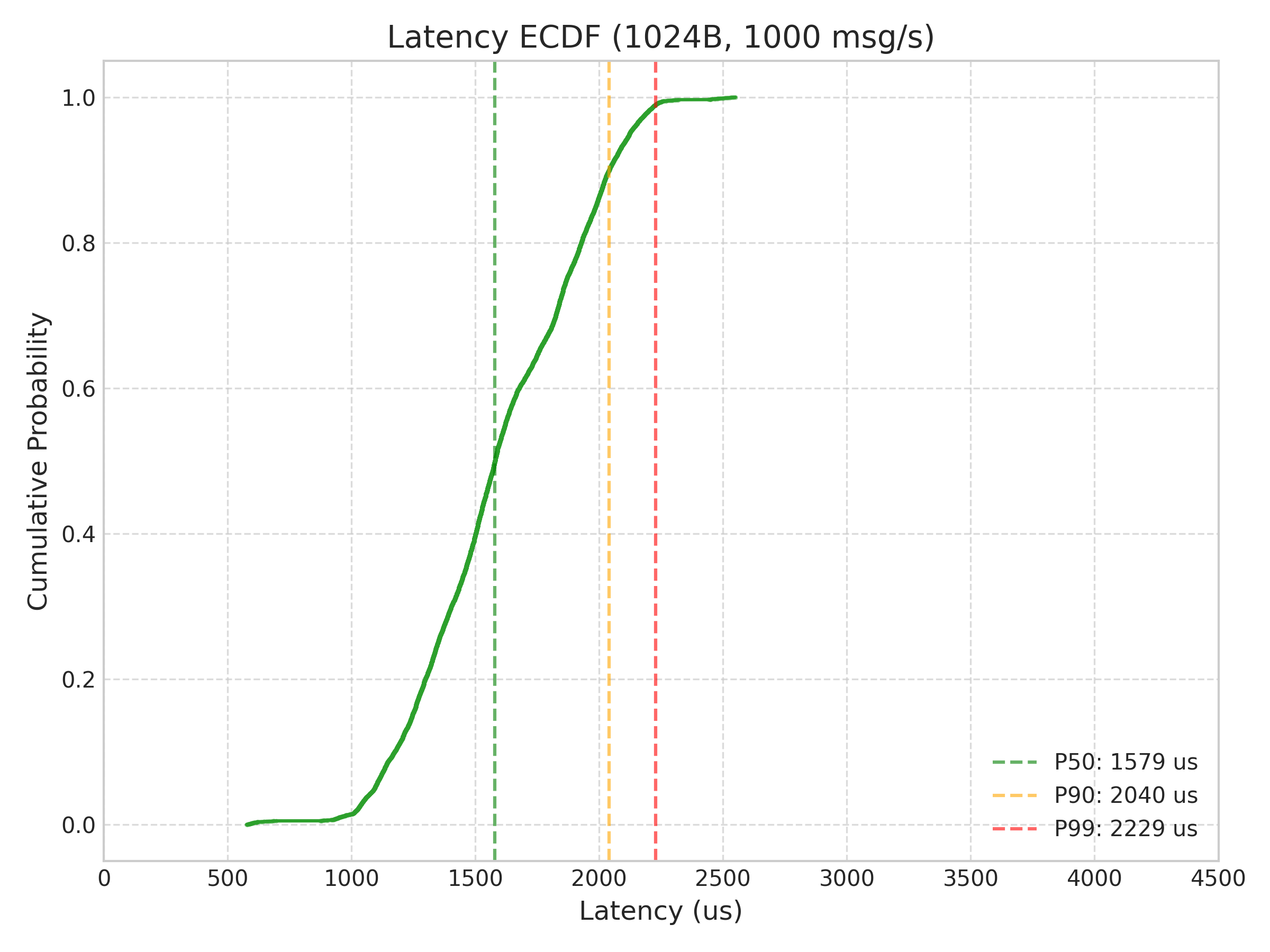}
  \caption{Payload=1024B, Rate=1000 msg/s.}
  \label{fig:bus-ecdf-1024-1k}
\end{subfigure}
\hfill
\begin{subfigure}[H]{0.48\linewidth}
  \centering
  \includegraphics[width=0.7\linewidth]{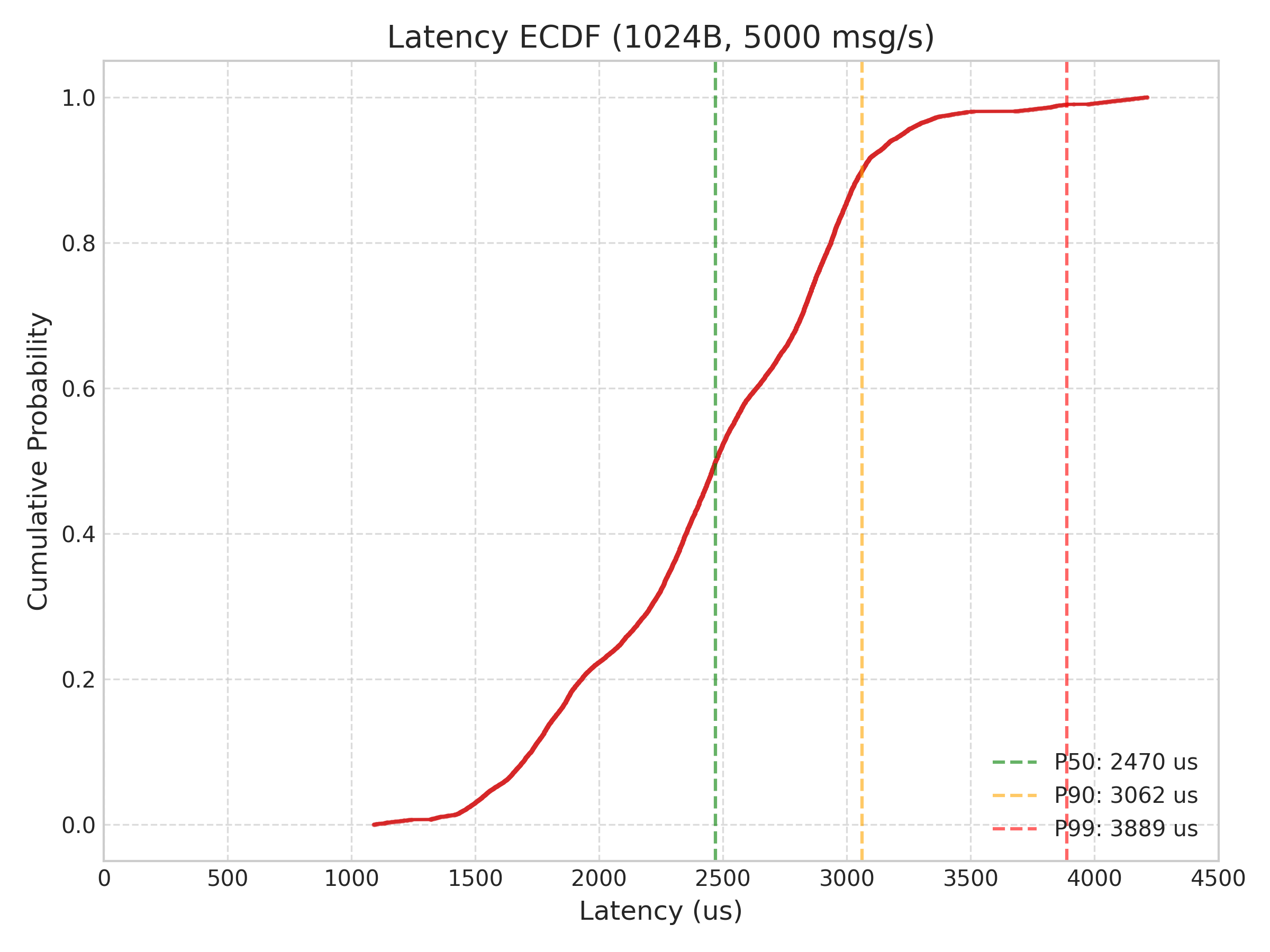}
  \caption{Payload=1024B, Rate=5000 msg/s.}
  \label{fig:bus-ecdf-1024-5k}
\end{subfigure}
\caption{Message-bus delivery latency ECDFs under a 1$\times$1 setup for four (payload size, publish rate) configurations.}
\label{fig:bus-ecdf}
\end{figure*}

At each cycle boundary, the policy module reads the latest state from \textit{canonical records}, generates the next action, and publishes it as a commands/actions message to the message bus.
Execution-side components subscribe to the corresponding commands/actions stream; upon receiving a command, they orchestrate remote endpoint invocations and execution steps, and then publish runtime state changes, anomalies, and outcomes as status/events back to the same message bus.
Designated subscribers consume these status/events signals and materialize selected feedback into \textit{canonical records}, providing updated context for subsequent decision making.
Because commands and feedback share the same message-bus path, closed-loop interactions can be strengthened incrementally without breaking component boundaries, while remaining traceable in deployment settings.

\begin{algorithm}[hbt]
\caption{Simplified upstream preprocessing and record update.}
\label{alg:preprocess}
\begin{algorithmic}[1]
\State \textbf{Input:} observation stream $\mathcal{S}$, window size $W$
\State \textbf{State:} window buffer $\mathcal{B}$, running aggregates $\mathbf{A}$
\For{each observation $o \in \mathcal{S}$}
  \State $x \leftarrow \textsc{Parse}(o)$
  \State $x \leftarrow \textsc{Clean}(x)$
  \State $f \leftarrow \textsc{Normalize}(x)$ \Comment{map to canonical features}
  \State $\mathcal{B}.\textsc{Append}(f)$
  \If{$\mathcal{B}$ spans window $W$}
    \State $\mathbf{A} \leftarrow \textsc{Aggregate}(\mathcal{B})$
    \State $\textsc{WriteRecords}(\mathbf{A})$ \Comment{update Canonical Records}
    \State $\mathcal{B}.\textsc{EvictOld}()$
  \EndIf
\EndFor
\end{algorithmic}
\end{algorithm}

The inference module is organized around projects.
Upstream inputs are grouped into project-specific inputs, and each project is pre-configured with its inference policy at deployment time, including the selected model and associated inference parameters.
At runtime, the inference module reads the latest project state from \textit{canonical records} and runs inference according to the configuration bound to that project to produce an action command; the command is then published to the message bus for downstream execution.

Execution modules subscribe to the commands/actions channel and wait for new commands in an event-driven manner.
Once a command arrives, an executor parses the command payload and triggers the corresponding execution logic.
During execution, the executor organizes key state transitions (e.g., start, success, or failure) and relevant runtime information into feedback events; after the action completes, it publishes the feedback as a status/events message back to the bus.
Designated subscribers consume these feedback messages and materialize selected signals into \textit{canonical records}.
Algorithm~\ref{alg:execute-feedback} shows simplified pseudocode of subscription, consumption, feedback publication, and record updates on the execution side.

\begin{algorithm}[t]
\caption{Simplified execution-side subscription and feedback integration.}
\label{alg:execute-feedback}
\begin{algorithmic}[1]
\State \textbf{Input:} commands/actions channel $\mathcal{C}$ on the message bus
\Loop
  \State $c \leftarrow \textsc{SubscribeAndReceive}(\mathcal{C})$
  \State $\textsc{Record}(c)$
  \State $r \leftarrow \textsc{Execute}(c)$ \Comment{$r$ contains status/outcome}
  \State $e \leftarrow \textsc{PackEvent}(r)$
  \State $\textsc{Publish}(e)$ \Comment{publish to status/events channel}
  \State $\textsc{Record}(e)$
  \State $\textsc{UpdateRecords}(e)$ \Comment{materialize feedback into Canonical Records}
\EndLoop
\end{algorithmic}
\end{algorithm}

With execution feedback materialized in \textit{canonical records}, subsequent inference steps observe updated context and issue new actions, completing the closed loop.

\subsection{Message Bus Latency Characterization}
\label{subsec:eval-bus}

We characterize message-delivery latency under a single-publisher/single-subscriber setup.
We vary payload size and publish rate and summarize delivery latency in a distributional form using ECDFs and percentiles.
Table~\ref{tab:bus-percentiles} reports P50, P90, and P99 for four representative (size, rate) configurations, and Fig.~\ref{fig:bus-ecdf} shows the corresponding ECDF curves.
Overall, increasing payload size or publish rate shifts the latency distribution to the right; meanwhile, within the tested load range, P99 remains at the millisecond scale and we do not observe a pronounced runaway long tail.

\begin{table}[t]
\centering
\caption{Message-bus delivery latency percentiles under a 1$\times$1 setup. Latencies are in $\mu$s.}
\label{tab:bus-percentiles}
\begin{tabular}{rrccc}
\toprule
Payload (B) & Rate (msg/s) & P50 & P90 & P99 \\
\midrule
128  & 1000 & 627.5  & 1333.5 & 1487.0 \\
128  & 5000 & 1322.5 & 1746.0 & 1840.0 \\
1024 & 1000 & 1363.0 & 1828.0 & 1941.0 \\
1024 & 5000 & 1901.0 & 2392.2 & 2781.1 \\
\bottomrule
\end{tabular}
\end{table}

\subsection{Recovery Under Service Restarts}
\label{subsec:eval-recovery}

\begin{figure*}[h]
  \centering
  \includegraphics[width=0.5\linewidth]{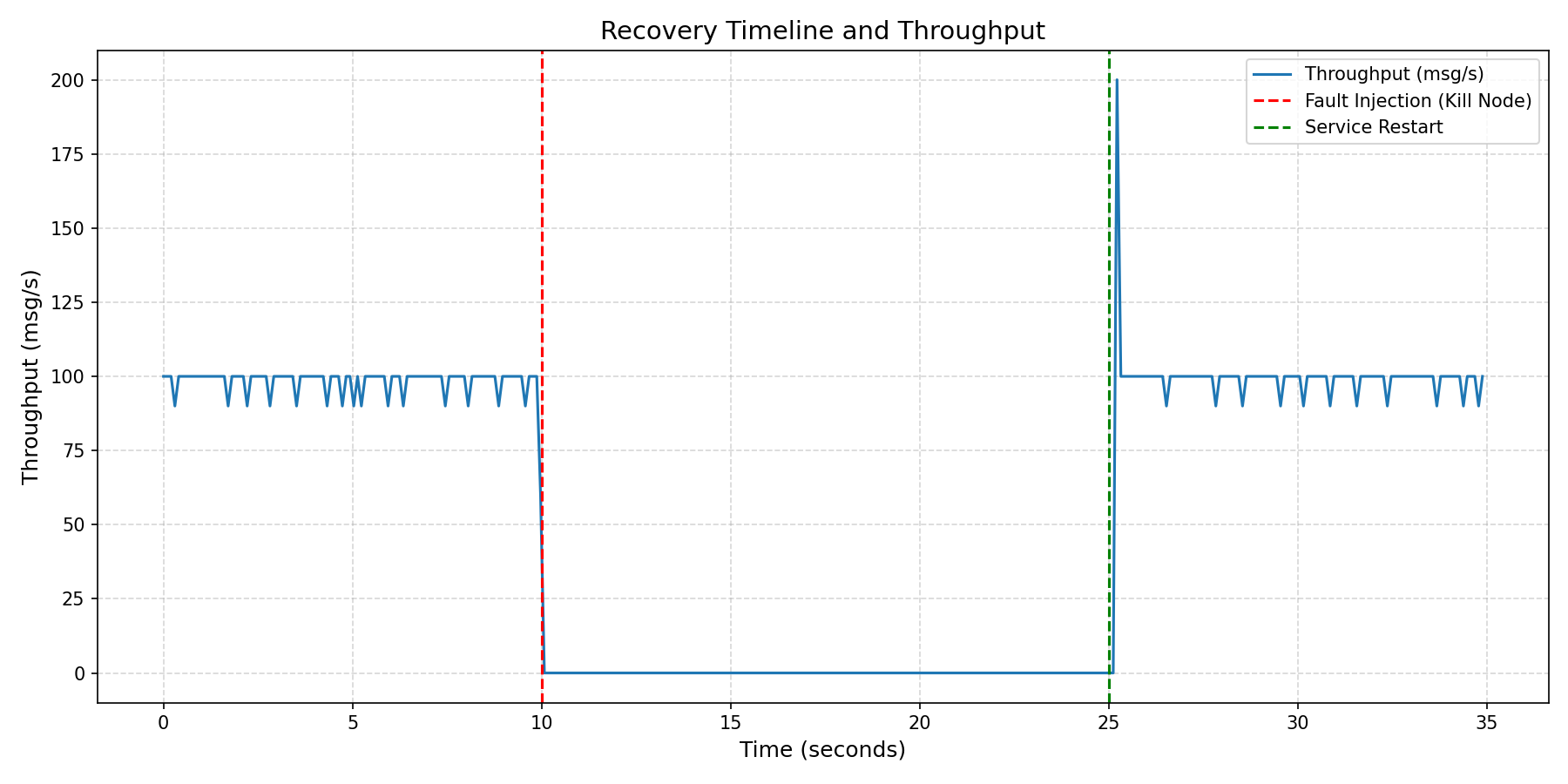}
  \caption{Delivered throughput over time under a controlled hard crash and restart. }
  \label{fig:recovery-throughput}
\end{figure*}

We evaluate recoverability via a controlled crash-and-restart experiment under a hard-crash setting.
We use a single-publisher/single-subscriber setup: the sender continuously publishes fixed-payload messages at a constant rate, and the receiver measures delivered throughput using fixed time bins.
The experiment begins in a steady regime where throughput stays close to the target rate.
We then inject a fault on the service node by force-killing the process and removing the shared-memory segment to emulate a cold restart; delivered throughput immediately drops to zero, indicating a disrupted stream.
After an intentionally configured downtime window, we restart the service node.
As the service becomes available again, clients trigger reconnection and rebuild the communication path, and delivered throughput returns to the steady regime with a brief transient around the recovery point.
Fig.~\ref{fig:recovery-throughput} summarizes the full trace.
The result indicates that, even under a worst-case hard crash with shared-memory loss in our testbed, the system can automatically resume delivery after the service comes back online without manual intervention.

\section{Mechanism Analysis}
\label{sec:analysis}

This section analyzes the mechanisms behind the latency distributions and recovery behavior observed in Sec.~\ref{sec:eval}.
We focus on two critical paths: (i) the message-bus delivery path, which explains how latency distributions shift as payload size and publish rate increase, and
(ii) the connection-management path, which explains how the system resumes delivery after a service crash and restart.
Our goal is to provide a concise, reproducible analysis grounded in observable runtime behaviors, without introducing new comparative baselines.

For the message-bus delivery path, end-to-end delivery latency can be decomposed into three stages:
publisher-side preparation and enqueueing, service-side routing and buffer handling, and subscriber-side dequeueing and dispatch.
Increasing payload size raises per-message costs (e.g., payload handling and buffer management), while increasing publish rate increases queue occupancy and contention.
Together, these effects produce the distribution-wide right shift observed in Sec.~\ref{subsec:eval-bus} and the consistent increase of P50/P90/P99 across configurations.

The ECDF curves in Sec.~\ref{subsec:eval-bus} further suggest that higher load primarily introduces additional processing and waiting time rather than sporadic extreme outliers.
When publish rate increases, the system spends a larger fraction of time in non-empty-queue regimes, and percentile growth reflects accumulated queueing delay.
When payload size increases, both the service and clients incur higher per-message handling costs, shifting the entire distribution.
Within the tested regime, this behavior is consistent with \emph{controlled degradation} under load rather than a qualitative runaway long tail.

Recovery, in contrast, is governed by control flow rather than steady-state queueing.
Fig.~\ref{fig:recovery-throughput} shows a sharp transition from steady delivery to zero throughput after fault injection, followed by resumption after service restart.
This trace indicates that the system can rebuild connectivity and re-establish subscriptions automatically after a hard crash with shared-memory loss, returning to a steady delivery regime without manual intervention (Sec.~\ref{subsec:eval-recovery}).

Mechanistically, the recovery procedure can be abstracted into three stages.
First, a \emph{fault-detection} stage triggers recovery upon error signals (e.g., connection errors, prolonged silence, or heartbeat timeouts).
Second, a \emph{cleanup-and-reconnect} stage closes stale connections, cancels pending operations, and applies a delayed reconnect window to avoid flapping under unstable conditions.
Third, a \emph{re-registration} stage re-advertises client identity and subscription information after connectivity is re-established so that routing state can be reconstructed and delivery can resume.
This staged view explains the end-to-end resumption pattern in Fig.~\ref{fig:recovery-throughput}.

In summary, Sec.~\ref{sec:eval} exhibits two complementary runtime properties:
(i) under increasing payload and publish rate, delivery latency grows in a controlled, distribution-wide manner consistent with load-induced processing and queueing costs; and
(ii) under a worst-case service crash with shared-memory loss, the system resumes steady delivery after restart through an automated recovery control flow.
These observations motivate the parameterization and optimization directions discussed next.

\section{Future Work}
\label{sec:future-work}

ANCHOR targets modularity and robustness at the system layer, but real-world deployments introduce additional requirements that remain outside the scope of this work.
We highlight two clusters of future directions: (i) security and privacy mechanisms that integrate with explicit shared state and inter-component communication, and
(ii) low-latency and real-time support under deployment constraints.

\subsection{Security and privacy for shared state and inter-component communication}

Real-world distributed, highly connected deployments turn shared context and message dissemination into key attack surfaces, requiring security and privacy to be treated as consistent system-level concerns across Canonical Records and the communication bus. This involves privacy-aware data handling for shared records with secure formats, encryption-at-rest and auditable access control for persisted and replayable scenarios \cite{tanimu2025addressing, durlik2024cybersecurity}; strengthened inter-component communication security via authenticated channels, least-privilege access control and blast-radius containment mechanisms \cite{durlik2024cybersecurity, hamad2023security}; integrated runtime intrusion detection, auditing and scalable verification for critical logic \cite{hamad2023security}; as well as post-quantum cryptography evaluation for long-lived systems to address classical crypto limitations \cite{durlik2024cybersecurity}.

\subsection{Low-latency and real-time support under deployment constraints}

The rising time-sensitivity of embodied tasks makes minimizing end-to-end and tail latency a priority, with the core challenge of preserving security and robustness while meeting real-time requirements. Future work should explore integration with real-time OS and kernel mechanisms like \\ PREEMPT-RT to boost scheduling determinism and reduce latency for time-critical loops \cite{ye2023ros2}; develop adaptive QoS strategies for publish-subscribe communication to balance reliability and responsiveness under variable workloads \cite{jalil2023performance}; optimize middleware and IPC via zero-copy transfer and enhanced serialization to cut coordination overhead \cite{kwok2025hprm}; and leverage edge and in-network computing to move processing closer to data sources, reducing round-trip delays for fast closed-loop responses in distributed setups \cite{chinta2024edge}.

\section{Conclusion}
\label{sec:conclusion}

In this paper, we presented \textsc{ANCHOR}, a modular framework for embodied AI systems that targets predictable runtime behavior under rapid iteration and deployment-time disturbances.
\textsc{ANCHOR} organizes cross-component interaction around explicit system-level interface contracts:
Canonical Records provide a unified and evolvable representation of normalized observations and system context, while a publish--subscribe messaging substrate enables decoupled coordination and many-to-many dissemination, supporting closed-loop execution and traceability without introducing tight coupling.

We empirically evaluated \textsc{ANCHOR} using an event-driven closed-loop workflow that instantiates the full sensing-to-execution loop:
upstream preprocessing materializes normalized state into \textit{canonical records}, inference produces and publishes commands according to project-bound configurations,
and execution components subscribe, act, and publish runtime status/events that are selectively materialized back into the shared context.
This evaluation demonstrates how explicit representation contracts and message-mediated interaction preserve clear component boundaries as modules evolve, enabling traceable operation and iterative debugging.

We further characterized message-delivery latency under increasing load.
By varying payload size and publish rate, we observed a smooth distribution-wide right shift in latency as load increases,
with tail percentiles remaining at the millisecond scale within the tested regime and without exhibiting runaway long tails.
These results are consistent with the intended system behavior: controlled degradation under overload rather than non-linear latency spikes.

Finally, we evaluated recoverability under a controlled crash-and-restart setting.
A hard crash causes delivered throughput to drop to zero, and delivery resumes after restart without manual intervention.
This behavior follows from the recovery control flow for fault detection, cleanup, reconnection, and re-registration, and remains effective even under worst-case conditions in our testbed such as shared-memory loss (Fig.~\ref{fig:recovery-throughput}).

Looking forward, we will prioritize two directions.
First, strengthening safety and security mechanisms---including access control, data isolation, and end-to-end transport protection---to reduce risk when operating closed-loop systems in open environments.
Second, improving end-to-end responsiveness and determinism by optimizing critical paths and developing parameterized QoS and real-time support to further reduce tail latency under complex workloads.
We believe that extending safety and low-latency properties on top of a long-lived modular framework such as \textsc{ANCHOR} is a practical path toward scalable and robust embodied-AI deployment.

\bibliographystyle{elsarticle-num} 
\bibliography{library}

\end{document}